\documentclass{WileyMSP-template}
\usepackage[numbers,sort&compress]{natbib}
\usepackage{amsmath}
\usepackage{indentfirst}
\usepackage{amsfonts,amssymb}
\usepackage{cases}
\usepackage{algorithm,algorithmic}
\usepackage{ragged2e}
\usepackage[hyphens]{url}
\usepackage{breakurl}
\usepackage{setspace}

\begin{document}

\pagestyle{plain}
\fancyhf{}

\noindent \title{A Physics-driven GraphSAGE Method for Physical Process Simulations Described by Partial Differential Equations}

\maketitle

\author{Hang Hu}
\author{Sidi Wu}
\author{Guoxiong Cai}
\author{Na Liu*}

\begin{affiliations}
\noindent Hang Hu, Guoxiong Cai, Na liu\\
\noindent Institute of Electromagnetics and Acoustics\\
\noindent Xiamen University\\
\noindent 4221th Xiang'an South Road, Xiang'an District, Xiamen, 361102, China\\

\noindent Email Address: hangh@stu.xmu.edu.cn; gxcai8303@xmu.edu.cn; liuna@xmu.edu.cn\\
\begin{spacing}{1.2}
\end{spacing}
\noindent Sidi Wu\\

\noindent School of Mathematical Sciences\\
\noindent Peking University\\
\noindent 5th Yiheyuan Road, Haidian District, Beijing, 100871, China\\
\noindent Email Address: wsd@pku.edu.cn
\end{affiliations}


\justifying

\begin{abstract}

Physics-informed neural networks (PINNs) have successfully addressed various computational physics problems based on partial differential equations (PDEs). However, while tackling issues related to irregularities like singularities and oscillations, trained solutions usually suffer low accuracy. In addition, most current works only offer the trained solution for predetermined input parameters. If any change occurs in input parameters, transfer learning or retraining is required, and traditional numerical techniques also need an independent simulation. In this work, a physics-driven GraphSAGE approach (PD-GraphSAGE) based on the Galerkin method and piecewise polynomial nodal basis functions is presented to solve computational problems governed by irregular PDEs and to develop parametric PDE surrogate models. This approach employs graph representations of physical domains, thereby reducing the demands for evaluated points due to local refinement. A distance-related edge feature and a feature mapping strategy are devised to help training and convergence for singularity and oscillation situations, respectively. The merits of the proposed method are demonstrated through a couple of cases. Moreover, the robust PDE surrogate model for heat conduction problems parameterized by the Gaussian random field source is successfully established, which not only provides the solution accurately but is several times faster than the finite element method in our experiments.
\end{abstract}

\section{Introduction}
Partial differential equations (PDEs) are an important category of mathematical models utilized in science and engineering. Several traditional numerical methods have been developed, including the finite element method (FEM), the finite difference method, and the finite volume method. Over the past few decades, these methods have undergone mature refinement, resulting in efficient implementations and mathematics theory guarantees. Inspired by successes in fields such as computer vision, speech recognition, and natural language processing, deep neural networks (DNNs) have recently been adopted as solvers for PDE-constrained computational problems.\textsuperscript{\cite{ren-qiang-2022-fourier-operator-learning,bao-jie-2020-couple-modeling-redox-flow-batteries,sato-rikuya-2023-machine-learning-semiconductor-described-by-couple-pde,zheng-si-qian-2019-numerial-study-thermal-optical-machine-learning}} Based on the way of constructing loss functions, these DNN solvers are classified into two categories: data-driven neural networks and physics-informed neural networks (PINNs).\textsuperscript{\cite{G.E.-2019-PINN}}

Over the past few years, data-driven neural networks have seen extensive applications in developing parametric surrogate models, demonstrating excellent performance, including acceptable accuracy and rapid inference.\textsuperscript{\cite{peng-jiang-zhou-2022-reduced-order,e.prume-2023-data-driven-mechanics,lin-jiang-peng-2022-thermal-prediction}} However, the training of data-driven approaches relies on vast amounts of labeled data, which necessitates substantial traditional numerical simulations to generate, resulting in large time expenditures. Furthermore, data-driven models are vulnerable to noisy data and exhibit poor generalization due to exclusive training on labeled data. Recently, PINNs have attracted considerable attention by reducing the demand for vast amounts of labeled data. The loss function of a PINN comprises a weighted combination of several terms, including mismatches between predicted solutions and labeled data, residuals of governing equations, and residuals of boundary conditions. Based on the point-wise formulation of the loss function, automatic differentiation\textsuperscript{\cite{baydin-2017-automatic-differential}} is utilized to compute differential terms in PDEs. Consequently, PINNs exhibit resilience to noisy data and strong generalization capabilities as their training focuses on PDEs rather than labeled data. They have found widespread applications in the field of numerical computation, demonstrating impressive performance in high-dimensional problems,\textsuperscript{\cite{wang-yi-ran-2024-extreme-learning-machine-high-dimension,zang-yao-hua-2020-weak-adversarial-networks-high-dimension}} nonlinear problems,\textsuperscript{\cite{yuan-lei-2022-A-pinn-nonlinear,tang-si-ping-2023-polynomial-interpolation-nonlinear}} and inverse problems.\textsuperscript{\cite{ameya-d-2020-cpinn-inverse-problem,ameya-d-2022-xpinn-inverse-problem}}

Despite promising prospects of PINNs, some challenges remain unsettled, particularly in solving PDEs with irregular solutions (e.g., singularities, oscillations) and developing PDE parametric surrogate models for rapid inference.\textsuperscript{\cite{hu-tian-hao-2023-Elliptic-Problems-with-Singular-Sources,Zhang-rui-2023-high-frequency,sun-yu-biao-2023-pinn-airfoil-geometry}} Neural networks inherently approximate smooth solutions over unsmooth ones. While applying vanilla PINNs to PDEs involving singularities (e.g., geometric, source singularities), special treatments are necessary for accurately obtaining sharp solutions. For example, Hu et al.\textsuperscript{\cite{hu-tian-hao-2023-singularity}} partitioned PDEs with singularities into two components: a regular part and a singular part, with the latter being represented by Fourier series for easier approximation. Yadav et al.\textsuperscript{\cite{Yadav-2022-distributed-pinn-singularity}} employed a distributed PINN that divided the computational domain into several sections to solve problems in solid mechanics with singularities, and Zeng et al.\textsuperscript{\cite{zeng-2024-adaptive-singular-problems}} proposed an adaptive (loss function/activation function/sampling method) PINN method for problems with corner singularities. In the aforementioned works, these specially designed approaches have demonstrated competent performance in addressing singularities. However, they relied on point-wise loss functions, thereby facing challenges in tuning penalty coefficients when applying soft boundary constraints.\textsuperscript{\cite{wang-sifan-2021-soft-constrain,WU-sidi-2022-soft-constrain}} Another form of irregular solution shows the oscillatory behavior resulting from high-frequency PDEs. Neural networks prefer to approximate functions at lower frequencies over higher ones, which is referred to as the Frequency Principle.\textsuperscript{\cite{xu-2020-F-principle}} To settle this challenge, various methods have been proposed, including, but not limited to, high-order deep neural networks (HOrderDNN),\textsuperscript{\cite{chang-2022-high-order-dNN-high-frequency}} finite basis physics-informed neural networks (FBPINNs),\textsuperscript{\cite{moseley-2023-FBPINN-domian-decomposition}} multi-scale deep neural network (MscaleDNN)\textsuperscript{\cite{liu-2020-multiscaleNN}} and phase shift deep neural networks (PhaseDNN).\textsuperscript{\cite{cai-2020-phaseDNN-high-frequency}} Specifically, HOrderDNN introduced a high-order nonlinear feature layer between the input layer and the first hidden layer to enhance approximation capabilities for spectral bias issues. In FBPINNs, the oscillatory PDE solution was considered a summation of finite basic functions learned by a low-capacity PINN. MscaleDNN combined radical scaling in the frequency domain with compactly supported activation functions to tackle the solution with multi-frequency content. PhaseDNN transformed high-frequency components into low-frequency ones for simpler approximation. Overall, the common strategy for handling PDEs with oscillatory behavior is constructing the high-frequency solution utilizing low-frequency components.

The process of solving PDEs is usually non-parametric, implying that any changes in input parameters necessitate retraining PINNs or applying transfer learning, lacking the instant inference capability characteristic of data-driven neural networks. However, numerous complicated engineering and computational physics problems require repetitive evaluations, including PDE-restricted optimization problems,\textsuperscript{\cite{rakhoon-hwang-2023-pde-constrained,karbasian-2022-shape-optimization}} uncertainty quantification,\textsuperscript{\cite{Rohit-K-Tripathy-2018-deepUQ, Kwon-Yongchan-2020-UQ-Bayesian-NN}} and topology optimization.\textsuperscript{\cite{hunter-t-kollmann-2020-topology-optimization, Chandrasekhar-2021-tounn}} PINNs based on convolutional neural networks (CNNs) and operator learning approaches are commonly employed to develop PDE parametric surrogate models without labeled data. For instance, Qi et al.\textsuperscript{\cite{QI-2023-FDTD-PINN}} developed a two-dimensional CNN-based parametric heat conduction surrogate model, coupled with a single-physics numerical solver for electromagnetic field simulation, to solve electromagnetic-thermal problems. Fuhg et al.\textsuperscript{\cite{Fuhg-2023-DEEPritz-parametric-pde}} combined the Ritz method with CNN-based PINN to construct parametric PDE surrogate models, while Koric et al.\textsuperscript{\cite{KORIC-2023-deeponet-heat-conduction}} evaluated the performance of data-driven DeepONet and physics-informed DeepONet methods in solving two-dimensional heat conduction equations with random source terms. Although CNN-based PINNs demonstrate noteworthy efficiency and scalability due to their effective convolution operations, they are limited to rectangular grids because of the finite difference theory used for numerical differentiation. This limitation results in high computational demand when refining is required in irregular areas. In contrast to CNN-based PINNs, operator learning is considerably more complex to implement due to its sophisticated structures. To model realistic and complicated PDE-governed problems cost-effectively, neural network solvers based on graph representations have recently gained significant attention since they can describe the computational domain with fewer evaluated points.\textsuperscript{\cite{hwang-2023-point-clouds,peng-jiang-zhou-2023-gnn-geometry-adaptive-natural-convection}} Gao et al.\textsuperscript{\cite{HanGao-2022-PI-GGN}} introduced a physics-informed graph convolutional network (GCN) framework based on the discrete weak form of PDEs to address both forward and inverse problems. However, this method is confined to the non-parametric setting. Graph Sample and Aggregate (GraphSAGE)\textsuperscript{\cite{William-L-Hamilton-2018-GraphSAGE}} is a spatial-based graph neural network that employs the neighbor sampling strategy, making it suitable for large-scale graph computation and offering lower time complexity compared to spectral-based GCNs.\textsuperscript{\cite{kipf-2017-GCN}} Nevertheless, it is usually used for data-driven training.

In this work, a new physics-driven GraphSAGE (PD-GraphSAGE) framework based on the weak form of PDEs discretized using piecewise polynomial nodal basis functions is proposed. Since the boundary conditions are strictly and naturally enforced, the turning of penalty coefficients is avoided. Moreover, flexible graph representations of the physical domain reduce the need for evaluated points. This approach aims to resolve computational physics problems exhibiting singularities or oscillations and to build parametric surrogate models for rapid and accurate inference of PDE solutions. Based on PD-GraphSAGE, our specific contributions are summarized as follows:
\begin{itemize}
    \item [1)]
    A distance-related coefficient is devised as the edge feature to solve PDEs with singularities;
    \item [2)]
    A feature mapping strategy is proposed to handle PDEs with oscillations;
    \item [3)]
    Real-time parametric PDE surrogate models are established for instant inference.
\end{itemize}

The structure of this paper is as follows: Section 2 provides details about the construction of PD-GraphSAGE, including the graph representation and computing method for solving computational physics problems, the message-passing principle of GraphSAGE, two proposed strategies, the derivation of the loss function using the Galerkin method, and the enforcement of hard boundary constraints. Subsequently, the overall framework of the proposed method is elaborated. In section 3, we investigate five distinct physical simulations, including electrostatics field problems, electromagnetics problems, and parametric steady heat conduction problems, to demonstrate the effectiveness of the proposed method. Section 4 offers conclusions and outlooks drawn from the entire work.

\section{Methodology}
\subsection{Architecture of PD-GraphSAGE}
\subsubsection{Graph Representations and Computing for Problems in Computational Physics}
\begin{figure}[htbp]
    \centering
    \includegraphics[width=0.75\linewidth]{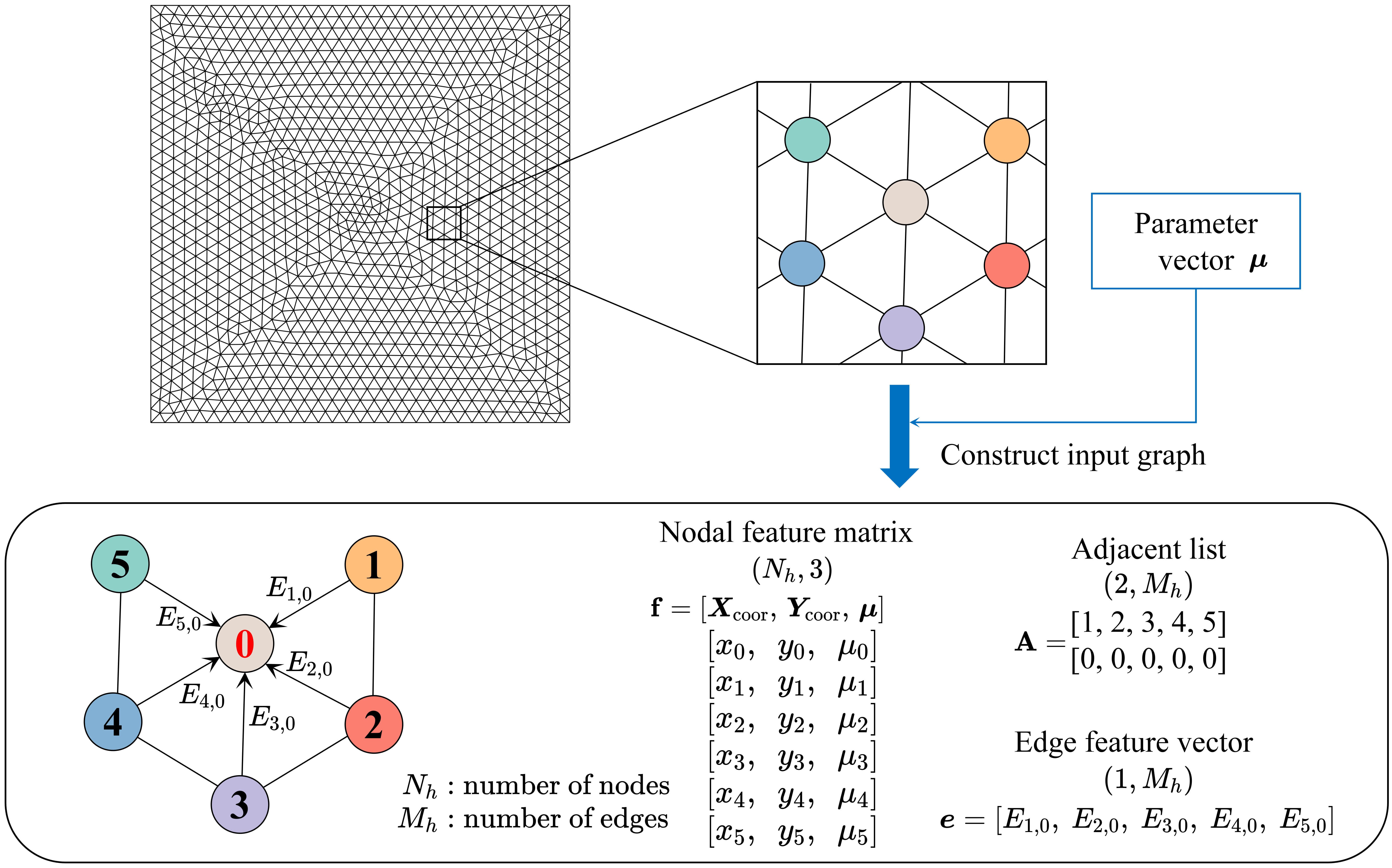}
    \caption{The unstructured grid is represented as a graph (node 0 is the target node for simplicity). Columns in the nodal feature matrix $\mathbf{f}$ of the input graph represent nodal $x$-coordinates $\boldsymbol{X}_\text{coor}$, $y$-coordinates $\boldsymbol{Y}_\text{coor}$, and the input parameter vector $\boldsymbol{\mu}$ (e.g. the source terms), respectively. Each item in the edges feature vector $\boldsymbol{e}$ is a distance-related value. Each column in the adjacent list $\mathbf{A}$ denotes the directional connection between nodes of the first row to the second row.}
    \label{fig:graph-example}
\end{figure}

Graph Neural Networks (GNNs) play an increasingly significant role in various scientific fields,\textsuperscript{\cite{Roberto-Perera-2023-gnn-displacement-crack-filds, David-Dalton-2023-gnn-soft-tissue, peng-jiang-zhou-2023-gnn-fluid-flow}} thanks to their powerful ability to handle highly expressive graph data. In the process of FEM implementation, triangular elements are used to accurately describe the computational domain, which is highly consistent with the graph format. As a two-dimensional example shown in \textbf{Figure \ref{fig:graph-example}}, vertex coordinates matrix $\mathbf{x}=[\boldsymbol{X}_\text{coor},\boldsymbol{Y}_\text{coor}]$, the adjacent list $\mathbf{A}$ and edge feature vector $\boldsymbol{e}$ can be extracted from the discrete mesh which is locally refined when needed, and are combined with the parameter vector $\boldsymbol{\mu}$ (e.g. source terms) to construct the input graph $\mathbf{G}(\mathbf{f},\boldsymbol{e},\mathbf{A})$ of the GNN, where $\mathbf{f}=[\boldsymbol{X}_\text{coor},\boldsymbol{Y}_\text{coor},\boldsymbol{\mu}]$ is the input nodal feature matrix. The GNN takes $\mathbf{G}$ as the input and outputs the graph $\mathbf{G}^{'}(\boldsymbol{f}^\text{out},\boldsymbol{e},\mathbf{A})$, where $\boldsymbol{f}^{\text{out}}$ is the predicted nodal solution-related values and is usually used to calculate the mean square error with labeled data to drive the training of the network. In this work, we utilize governing PDEs instead of tremendous amounts of labeled data to train the GNN.

\subsubsection{Chosen Basic Framework of GNN}
\begin{figure}[htbp]
    \centering
    \includegraphics[width=0.75\linewidth]{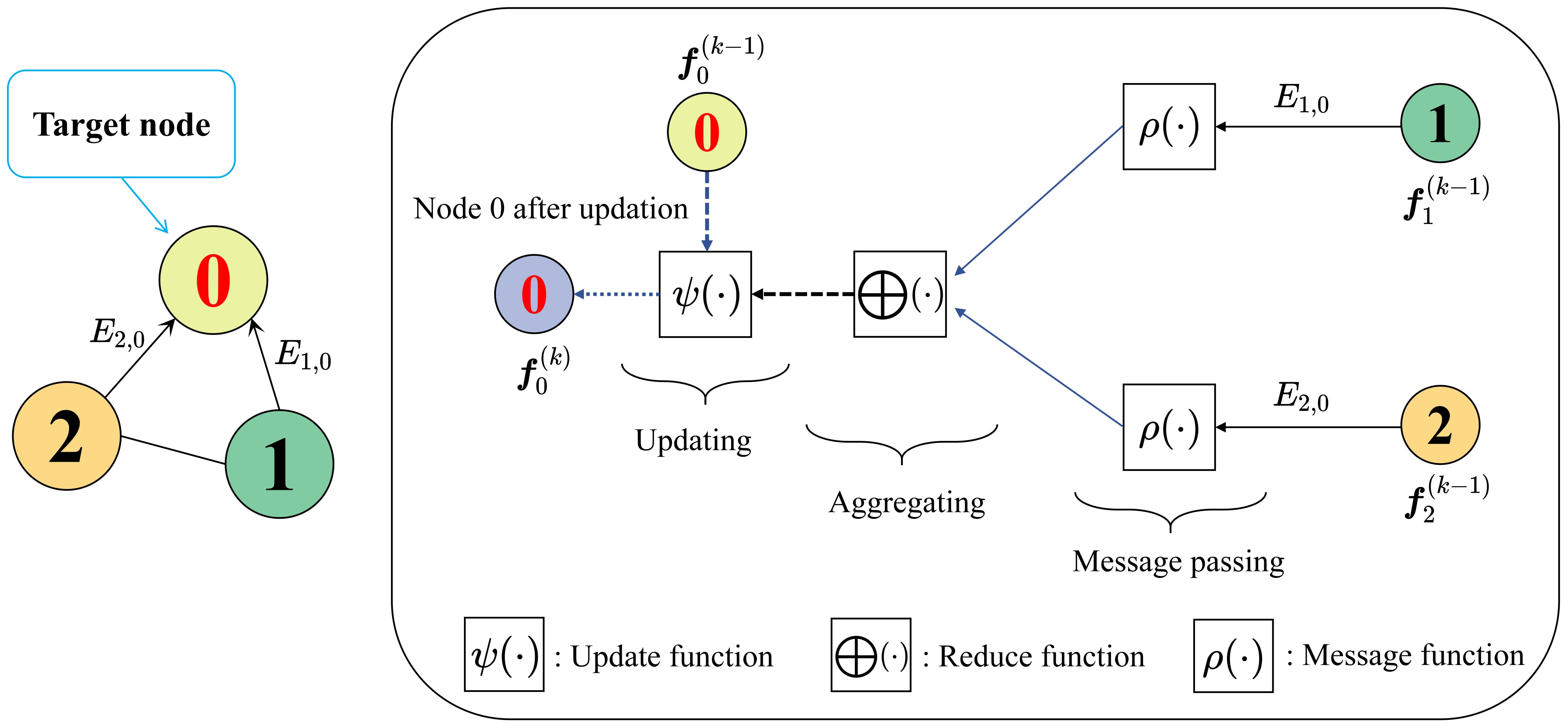}
    \caption{An example of updating nodal features of node 0 in the $k$th GNN layer. It involves three steps: message passing, message aggregation, and feature updating, enabling the target node 0 to update its features from $\boldsymbol{f}_{0}^{(k-1)}$ to $\boldsymbol{f}_{0}^{(k)}$.}
    \label{fig:graph-process}
\end{figure}

As illustrated in \textbf{Figure \ref{fig:graph-process}}, the essence of GNN lies in the message-passing paradigm.\textsuperscript{\cite{Justin-Dahl-2017-mpnn}} This process can be briefly described as: 
\begin{align}
	 \boldsymbol f_i^{(k)}=\psi^{(k)} \left(\boldsymbol f_i^{(k-1)}, \bigoplus\limits{_{j \in \mathcal{N}{(i)}}^{(k)}}\rho^{(k)} \left(\boldsymbol f_i^{(k-1)},\boldsymbol f_j^{(k-1)},E_{j,i}^{(k-1)}  \right)   \right)\quad(k\geq1,\ \boldsymbol{f}_i^{(0)}=\left[{X}_\text{coor},{Y}_\text{coor},{\mu}\right]^\top),
\end{align}
where $\boldsymbol f_{i}^{(k)}$ and $\boldsymbol f_{j}^{(k-1)}$ represent features of the target node $i$ and the source node $j$ at $k$th and $(k-1)$th GNN blocks respectively, $\mathcal{N}{(i)}$ indicates the set of nodes directly connected to node $i$, and $E_{j,i}^{(k-1)}$ denotes the edge feature from node $j$ to node $i$. The message function $\rho(\cdot)$ and update function $\psi(\cdot)$ are both non-linear and differentiable, while $\bigoplus(\cdot)$ represents a differentiable, permutation-invariant reduce function such as summation, averaging, or taking the maximum.

In this work, we employ GraphSAGE neural network as the foundational framework. GraphSAGE is the precursor to PinSAGE, renowned as the inaugural graph-based algorithm effectively implemented in industrial-scale recommendation systems.\textsuperscript{\cite{Ying-2018-PinSAGE}} GraphSAGE operates in the spatial domain, exhibiting simpler time complexity compared to frameworks functioning in the spectral domain, such as GCN. Moreover, this approach employs a neighbor sampling strategy, resulting in reduced memory consumption during training and enhanced generalization capability. This makes it particularly well-suited for processing large-scale graphs, a common situation in modeling complex and realistic problems. The message-passing mechanism of GraphSAGE is formulated as follows:
\begin{align}
\boldsymbol f_{\mathcal{N}(i)}^{(k)} &= \text{mean}\left(\{E_{j,i}^{(k-1)} \boldsymbol f_{j}^{(k-1)}, \forall j \in \mathcal{N}(i) \}\right),
\\
\boldsymbol f_{i}^{(k)} &= \sigma^{(k-1)}\left( \mathbf{W}^{(k-1)}_1 
\left[\boldsymbol f_{i}^{(k-1)},\boldsymbol f_{\mathcal{N}(i)}^{(k)}\right]\boldsymbol{W}_2^{(k-1)} 
+\boldsymbol{b}^{(k-1)} \right),
\end{align}
where $\boldsymbol f_{\mathcal{N}(i)}^{(k)}$ are the aggregated nodal features, calculated as the mean values of $\boldsymbol{f}_j^{(k-1)}$, with each weighted by the corresponding predefined edge feature $E_{j,i}^{(k-1)}$. $\mathbf{W}_1^{(k-1)}$, $\boldsymbol{W}_2^{(k-1)}$, and  $\boldsymbol{b}^{(k-1)}$ are the GNN block parameters to train, and $\sigma^{(k-1)}$ is the non-linear activation function.

\subsubsection{Distance-related Edge Feature and Feature Mapping Strategies}
In this part, we put forward two newly useful strategies to promote the training of the GNN. One of the important components of a graph is the edge feature. Adopting the proper edge feature can greatly facilitate the convergence of the GNN.\textsuperscript{\cite{shi-2022-feature-attention,meng-2020-enhance-edge-feature}} In this work, we devise and consider the distance-related coefficient as the edge feature,  
\begin{align}\label{edgefeature}
E_{j,i}^d=\text{log}_{\epsilon}(\frac{L_\text{max}}{L_{j,i}}),
\end{align}
where $L_{j,i}$ is the physical distance between nodes $j$ and $i$, and $L_\text{max}$ is a self-defined scale factor larger than the distance between any two adjacent nodes. It can be easily deduced that if the distance between two nodes is smaller, their solution values are more similar. Therefore, $1/L_{j,i}$ appears to be a suitable edge feature as a weight to foster the training of the GNN. However, when $L_{j,i}$ is extremely small, the weight becomes unexpectedly large, leading to unstable training. Consequently, $L_\text{max}$ is set to ensure $L_\text{max}/L_{j,i}\textgreater1$, and $\text{log}_{\epsilon}(\cdot)$ acts as a compression function to prevent huge differences between weights, where $\epsilon$ is a number slightly greater than 1. If the proposed distance-related edge feature is used, $E_{j,i}:=E_{j,i}^d$ and pre-experimental results indicate that $E_{j,i}^d$ demonstrates optimal effectiveness when employed exclusively in the first hidden layer. In this work, ${\epsilon}$ and $L_\text{max}$ are set to 1.05 and 1 m as default respectively, and the parameter sensitivity analysis will be given in the following case 1. However, $E_{j,i}^d$ is not suitable for oscillation occasions, such as those defined by the high-frequency Helmholtz equations, because it deteriorates the approximation process of the locally rapidly fluctuating solution. In such cases, $E_{j,i}$ is set to 1 for all edges by default.

As issues defined by PDE with high frequencies are challenging for vanilla PINN to solve, Tancik et al.\textsuperscript{\cite{tancik-matthew-2020-fourier-feature-mapping}} presented a novel method called the Fourier feature mapping to overcome the spectral bias. The input nodal features $\mathbf{f}$ are lifted to multiscale dimensions composed of a series of sinusoids via mapping:
\begin{align}
\gamma_1(\mathbf f)=\left[\cos(2\pi\mathbf f\mathbf{B}),\sin(2\pi\mathbf f\mathbf B)\right],
\end{align}
where each item in the lift matrix $\mathbf B\in\mathbb{R}^{q\times m_1}$ ($q$ is the number of nodal features) is sampled from $N(0,\sigma_1 ^2)$. Here, drawing inspiration from this method, we devise another feature mapping strategy:
\begin{align}
\gamma_2(\mathbf f)=[\cos(\boldsymbol{C}\otimes\mathbf f),\sin(\boldsymbol{C}\otimes\mathbf f)],
\end{align}
where each item in the scale vector $\boldsymbol{C}\in\mathbb{R}^{1\times m_2}$ is sampled from $N(0,\sigma_2 ^2)$, and $\otimes(\cdot)$ is the kronecker product. In these two mappings, $m_i$ and $\sigma_i$  $(i=1, 2)$ are hyperparameters sensitive to the weights and biases initialization of the neural network, representing the number of nodal features between the input layer and the first hidden layer and the corresponding degree of feature dispersion respectively, which can be optimized via parameter sweeping.

\subsection{PDEs-driven Loss Function}
In this section, we will introduce the details of how to construct the loss function of the GNN based on PDEs and the way to implement boundary conditions with hard enforcement. Consider a general parametric PDE system of steady-state with order $O$ ($O\geq 2)$,
\begin{subnumcases}{}\label{eq:governing equation and boundary}
\nabla \cdot \boldsymbol{\mathcal{F}_1}\left(u;\boldsymbol{\mu}\right)+\mathcal{F}_2(u;\boldsymbol{\mu})=0&$\text{in}\ \Omega,$\\
\mathcal{B}(u;\boldsymbol{\mu})=0&$\text{on}\ \partial\Omega$,\label{6b}
\end{subnumcases}
which is defined on the physical domain $\Omega \subseteq \mathbb{R}^d\ (d= 2\ \text{or}\ 3)$, $u$ is the solution variable, $\boldsymbol{\mu}$ is the spatial location-related parameter vector, $\nabla:=(\partial_{x_1},\cdots,\partial_{x_d})$ is the gradient operator, $\boldsymbol{\mathcal{F}_1}:\mathbb{R}\rightarrow\mathbb{R}^{1\times d}$ and $\mathcal{F}_2:\mathbb{R}\rightarrow\mathbb{R}$ are the flux operator and source operator respectively, and $\mathcal{B}:\mathbb{R}\rightarrow\mathbb{R}$ is the boundary operator, which is implemented on the boundary $\partial\Omega$ of the physical domain.
\subsubsection{Applying Discrete Galerkin Method to PDEs}
The weak form of Equation \eqref{eq:governing equation and boundary} is obtained via the Galerkin method by multiplying the test function $w:\mathbb{R}^{1\times d}\rightarrow \mathbb{R}$ and integrating by part, which is
\begin{align}\label{weak form}
\int_{\partial\Omega}w\cdot \boldsymbol{\mathcal{F}_1}(u;\boldsymbol{\mu})\boldsymbol{n}\,\text{d}S-\int_\Omega  \nabla w:\boldsymbol{\mathcal{F}_1}(u;\boldsymbol{\mu})\,\text{d}V+\int_\Omega w \cdot \mathcal{F}_2(u;\boldsymbol{\mu})\,\text{d}V=0,
\end{align}
where $\boldsymbol{n}\in\mathbb{R}^d$ is the unit outward normal. Piecewise polynomial nodal basis functions $\boldsymbol{\phi}:\mathbb{R}^{1\times d}\rightarrow \mathbb{R}^{N_h}$ are introduced to construct an approximate representation of the solution $u$ on the discrete computational domain, where $N_h$ is the number of nodes. These functions are also used to express the test function as a basis in the Galerkin method, thus Equation \eqref{weak form} is turned to:
\begin{align}
\int_{\partial\Omega}\boldsymbol{\phi} \cdot\boldsymbol{\mathcal{F}_1}(u_h;\boldsymbol{\mu})\boldsymbol{n}\,\text{d}S-\int_\Omega  \nabla \boldsymbol{\phi}:\boldsymbol{\mathcal{F}_1}(u_h;\boldsymbol{\mu})\,\text{d}V+\int_\Omega \boldsymbol{\phi}  \cdot\mathcal{F}_2(u_h;\boldsymbol{\mu})\,\text{d}V=0.
\end{align}
As a result, the solving process of Equation (\ref{eq:governing equation and boundary}) and (7b)  can be viewed as a constrained optimization problem:
\begin{subequations}\label{eq:optimization}
\begin{align}
&\underset{\left[\boldsymbol{m}(\boldsymbol{\mu}_1),\cdots,\boldsymbol{m}(\boldsymbol{\mu}_N)\right]}{\operatorname*{ min}}\frac{1}{N}\sum_{i=1}^{N}\left\Vert\int_{\partial\Omega}
\boldsymbol\phi\cdot\boldsymbol{\mathcal{F}_1}(u_{h};\boldsymbol{\mu}_i)\boldsymbol{n}\,\text{d}S-\int_{\Omega}\nabla\boldsymbol\phi:\boldsymbol{\mathcal{F}_1}(u_{h};\boldsymbol{\mu}_i)\,\text{d}V+\int_{\Omega} \boldsymbol\phi\cdot\mathcal{F}_2(u_{h};\boldsymbol{\mu}_i)\,\text{d}V\right\Vert_2&\text{in}\ \Omega,\label{optimization1}\\
&\text{s.t.}\quad\mathcal{B}(u_h;\boldsymbol{\mu}_i)=0\quad(i=1,\cdots,N)&\text{on}\ \partial\Omega\label{optimization2},
\end{align}
\end{subequations}
where $N(\geq1)$ is the sampling setting number of parameter space to be trained, $\Vert \cdot \Vert_2$ denotes the 2-norm of the residual vector, the approximate numerical solution is $u_h(\boldsymbol{x},\boldsymbol{\mu}_i)=\boldsymbol{\phi}^\top(\boldsymbol{x})\boldsymbol{m}(\boldsymbol{\mu}_i)$, where the nodal coordinate $\boldsymbol{x}\in\Omega$ and the basis function coefficients $\boldsymbol{m}(\boldsymbol{\mu}_i)\in\mathbb{R}^{N_h}$ are the scaled output of GraphSAGE under specific parameter setting $\boldsymbol{\mu}_i\in\mathbb{R}^{N_h}$, i.e.,
\begin{align}
    {\boldsymbol{m}}(\boldsymbol{\mu}_i)= \delta{\boldsymbol{f}^\text{out}}(\boldsymbol{\mu}_i;\boldsymbol{\varTheta}),
\end{align}
where $\boldsymbol{\varTheta}$ is the training parameters of GraphSAGE, and $\delta$ is the empirical scale factor to regulate the scope of the solutions. By introducing the Gaussian integration points and weights $\left\{[\bar{\boldsymbol{x}}_j^s,\bar{\boldsymbol{x}}_k^v],[{{\alpha}}_j^s,{{\alpha}}_k^v]\right\}_{j=1,k=1}^{N_s,N_v}$ of the physical domain $\Omega$ and boundary $\partial\Omega$ ($N_s$ and $N_v$ is the number of Gaussian surface and volume integration points), each 2-norm residual item in Equation (\ref{eq:optimization}a) is converted to
\begin{align}\label{eq:residual}
\mathcal{R}\left({\boldsymbol{f}^\text{out}}\left(\boldsymbol{\mu}_i;\boldsymbol{\varTheta}\right);\boldsymbol{\mu}_i\right)
=&\sum_{j=1}^{N_{s}}\alpha_j^s \boldsymbol{\phi}(\bar{\boldsymbol{x}}_j^s)\cdot\boldsymbol{\mathcal{F}_1}\Big(\bar{u}_h(\bar{\boldsymbol{x}}_j^s,{\boldsymbol{f}^\text{out}}(\boldsymbol{\mu}_i;\boldsymbol{\varTheta}));\boldsymbol{\mu}_i\Big)\boldsymbol{n}\nonumber\\
&-\sum_{k=1}^{N_{v}}\alpha_k^v \nabla\boldsymbol{\phi}(\bar{\boldsymbol{x}}_k^v):\boldsymbol{\mathcal{F}_1}\Big(\bar{u}_h(\bar{\boldsymbol{x}}_k^v,{\boldsymbol{f}^\text{out}}(\boldsymbol{\mu}_i;\boldsymbol{\varTheta}));\boldsymbol{\mu}_i\Big)\\
&+\sum_{k=1}^{N_{v}}\alpha_k^v \boldsymbol{\phi}(\bar{\boldsymbol{x}}_k^v)\cdot\mathcal{F}_2\Big(\bar{u}_h(\bar{\boldsymbol{x}}_k^v,{\boldsymbol{f}^\text{out}}(\boldsymbol{\mu}_i;\boldsymbol{\varTheta}));\boldsymbol{\mu}_i\Big),\nonumber
\end{align}
where all the scalars and vectors are pre-calculated except the output ${\boldsymbol{f}^\text{out}}(\boldsymbol{\mu}_i;\boldsymbol{\varTheta})$ of GraphSAGE.
\subsubsection{Boundary Conditions Enforcement}
Boundary conditions described by Equation (\ref{6b}) can be categorized into natural boundary conditions and essential boundary conditions. In vanilla PINN, boundary conditions are softly applied, while some variants enforce them with simple algebraic expressions or small-capacity fully connected neural networks. This makes it challenging to strictly adhere to boundary conditions under complicated geometry situations. In this work, thanks to the ingenious construction and selection of piecewise polynomial nodal basis functions $\boldsymbol{\phi}$, natural boundary conditions are automatically satisfied in the surface integration term of the weak form, and essential boundary conditions can be explicitly encoded as follows:
\begin{align} 
{\boldsymbol{m}}(\boldsymbol{\mu}_i;\boldsymbol{\varTheta})=\begin{bmatrix}
\delta\boldsymbol{f}^\text{out}(\boldsymbol{\mu}_i;\boldsymbol{\varTheta})\\
{\boldsymbol{m}}_e(\boldsymbol{\mu}_i)\end{bmatrix}, 
\end{align}
where ${\boldsymbol{m}}_e(\boldsymbol{\mu}_i)$ represents the known essential boundary values.

Finally, from the standpoint of GraphSAGE training, our optimization target as illustrated in Equation (\ref{eq:optimization}), is to find the optimal GraphSAGE parameters $\boldsymbol{\varTheta}^*$ through the minimization of the loss function without labeled data:
\begin{align}
\boldsymbol{\varTheta}^*={\underset{\boldsymbol{\varTheta}}{\operatorname*{arg\,min}}\ \mathcal{L}_{}(\boldsymbol{\varTheta};\boldsymbol{\mu}_i)}\quad(i=1,\cdots,N).
\end{align}
In the context of parameterized problems, the objective is to develop parametric PDE surrogate models capable of instant inference, we define the loss function as: 
\begin{align}\label{loss function}
\mathcal{L}(\boldsymbol{\varTheta};\boldsymbol{\mu}_i)={\sum_{i=1}^{N}\Vert\mathcal{R}\left(\boldsymbol{f}^\text{out}(\boldsymbol{\mu}_i;\boldsymbol{\varTheta}),\boldsymbol{m}_e(\boldsymbol{\mu}_i);\boldsymbol{\mu}_i\right)\Vert_{2}}.
\end{align}
However, for non-parameterized problems characterized by a specific parameter setting (i.e. $N=1$), the loss function simplifies to: 
\begin{align}
\mathcal{L}(\boldsymbol{\varTheta};\boldsymbol{\mu})=\left\Vert  \mathcal{R}\left(\boldsymbol{f}^\text{out}(\boldsymbol{\mu};\boldsymbol{\varTheta}),\boldsymbol{m}_e;\boldsymbol{\mu}\right)\right\Vert_2,
\end{align}
which corresponds to scenarios where all instances of $\boldsymbol{\mu}$ are identical. Consequently, there is no need to incorporate the parameter vector $\boldsymbol{\mu}$ into the feature set of the input nodes $\mathbf{f}$.

\subsection{Uniform Framework}
\begin{figure}
\centering
\includegraphics[width=0.75\linewidth]{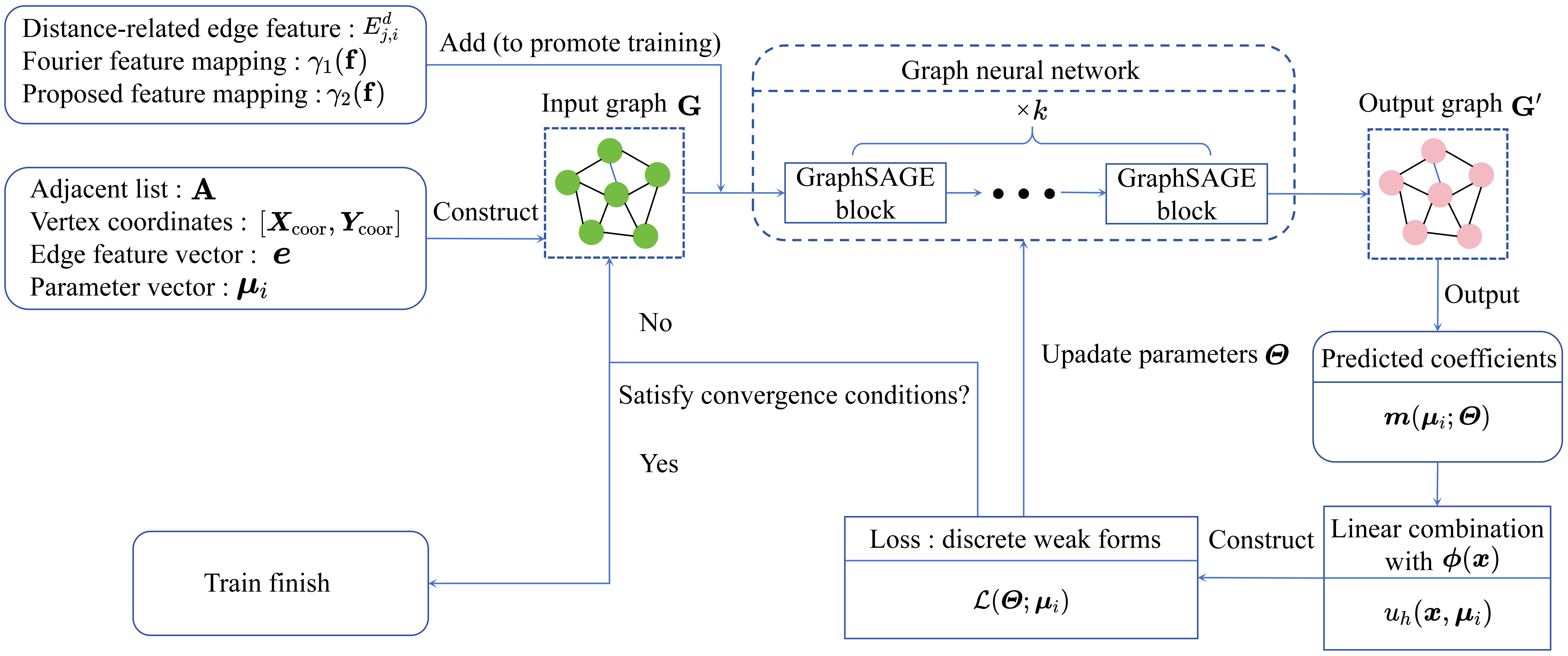}
\caption{The training schematics of the proposed PD-GraphSAGE.}
\label{fig:traing procedure}
\end{figure}

As depicted in \textbf{Figure \ref{fig:traing procedure}}, this section mainly outlines the proposed PD-GraphSAGE training scheme. Initially, the mesh information, including the adjacency list $\mathbf{A}$, vertex coordinates $\mathbf{x}=[\boldsymbol{X}_\text{coor},\boldsymbol{Y}_\text{coor}]$, edge feature vector $\boldsymbol{e}$, combined with the parameter vector $\boldsymbol{\mu}_i$, forms the input graph $\mathbf{G}$ for the network. Subsequently, the network consisting of $k$ GraphSAGE blocks employs feature mapping strategies $\gamma_1(\mathbf f)$ or $\gamma_2(\mathbf f)$ for oscillatory problems. In other scenarios, it utilizes the distance-related edge feature $E_{j,i}^d$ to enhance training convergence. The next step is linearly combining the output coefficients $\boldsymbol{m}(\boldsymbol{\mu}_i;\boldsymbol{\varTheta})$ with the piecewise polynomial nodal basis functions $\boldsymbol{\phi}$ to derive the nodal approximate solution ${u}_h(\boldsymbol{x},\boldsymbol{\mu}_i)$. Then, solutions are integrated into Equation (\ref{eq:residual}) and (\ref{loss function}) to formulate the loss function $\mathcal{L}(\boldsymbol{\varTheta};\boldsymbol{\mu}_i)$. Finally, the GNN iteratively updates its parameters $\boldsymbol{\varTheta}$ through loss backpropagation aiming at predefined convergence criteria, such as a specific number of epochs, a loss function threshold, a validation error threshold, and so on. Once the training is completed, inputting a new graph within the trained parameter space into PD-GraphSAGE enables it to quickly provide the solution. 

\section{Numerical Results}
This work presents five two-dimensional cases to evaluate the effectiveness of the proposed method in addressing PDE-constrained problems characterized by irregular solutions, and in constructing parametric PDE surrogate models. Case 1 and case 2 focus on electrostatics field problems defined by Poisson equations, where the singularities are caused by geometry and source, respectively. Case 3 and case 4 explore electromagnetic problems characterized by oscillatory solutions, defined by Helmholtz equations. Case 5 examines steady heat conduction problems defined by Poisson equations, parametrized by Gaussian random field heat sources.  All these cases are studied on a single NVIDIA GeForce RTX 3090 Ti-24G GPU and AMD Ryzen 5 5600x CPU cards, implemented on the open-source library Pytorch and DGL.\textsuperscript{\cite{wang-min-jie-2019-dgl}} Triangle elements and second-order piecewise polynomial nodal basis functions are used to discrete the computational domain and construct the approximate solution, $\text{sin}(\cdot)$ is used as the activation function throughout the network, Adam\textsuperscript{\cite{Diederik-2015-adam}} is used as the optimizer and learning rate is set as $1\times 10^{-4}$ for all cases. Coordinate inputs $\boldsymbol{x}$ are standardized to the range $[-1$ m,1 m] to boost the training, and the scale factor $\delta$ is set to 1 unless specifically stated. To verify the accuracy of the proposed framework, we define the relative error between the network prediction and numerical simulation results as: 
\begin{align}
\text{Relative }L^2 \text{ error} = \frac{\left\|{\boldsymbol{U}_\textrm{NN}} - {\boldsymbol{U}_\textrm{ref}} \right \|_2}{\left\|\boldsymbol{U}_\textrm{ref}\right \|_2},
\end{align}
where $\boldsymbol{U}_\textrm{NN}$ represents the numerical solution obtained by PD-GraphSAGE, and $\boldsymbol{U}_\textrm{ref}$ denotes the reference solution provided by FEM implemented on Matlab. Training is stopped when the (average) relative validation $L^2$ error is under $3\times 10^{-3}$ and no longer significantly decreases. Partial experimental settings for subsequent cases are summarized in \textbf{Table \ref{table:training details}}. Note that for the best performance, hyperparameters in the cases-$\gamma_1(\mathbf f)$-best ($\sigma_1$ and $m_1$) and the cases-$\gamma_2(\mathbf f)$ ($\sigma_2$ and $m_2$) are all acquired through parameter sweeping. However, in the cases-$\gamma_1(\mathbf f)$, hyperparameters are identical to those of the corresponding cases-$\gamma_2(\mathbf f)$ for direct comparison. 
\begin{table}[htbp]
	\centering
	\caption{Partial experiment configurations of PD-GraphSAGE.}
        \renewcommand\arraystretch{1.4}
	\setlength{\tabcolsep}{4mm}{
		\begin{tabular}{cccccc}
			\hline
			  &Layers& Edge feature& Feature mapping&Batch size& Convergent epochs\\
                \hline
			  Case 1&2+128$\times$5+1&$E_{j,i}^d$&   -  &1& $4\times 10^3$\\
                
                Case 2&2+256$\times$5+1&$E_{j,i}^d$&-  &1& $9.35\times 10^4$\\
                
                Case 3-$\gamma_1(\mathbf f)$-best& 2+20+256$\times$5+1& $E_{j,i}$=1& $\sigma_1=2$&1& $1.01\times 10^4$\\
 Case 3-$\gamma_1(\mathbf f)$& 2+40+256$\times$5+1& $E_{j,i}$=1& $\sigma_2=3$& 1&$1.06\times 10^4$\\
                 
                 Case 3-$\gamma_2(\mathbf f)$& 2+40+256$\times$5+1& $E_{j,i}$=1& $\sigma_2=3$&1& $1.48\times 10^4$\\
                 
                 Case 4-$\gamma_1(\mathbf f)$-best& 2+50+256$\times$5+1& $E_{j,i}$=1& $\sigma_1=1$&1& $5.58\times 10^4$\\
 Case 4-$\gamma_1(\mathbf f)$& 2+40+256$\times$5+1& $E_{j,i}$=1& $\sigma_2=12$& 1&$2.787\times 10^5$\\
                 
                 Case 4-$\gamma_2(\mathbf f)$& 2+40+256$\times$5+1& $E_{j,i}$=1& $\sigma_2=12$&1& $9.9\times 10^4$\\
                 
                 Case 5-$r=6$& 3+512$\times$5+1& $E_{j,i}^d$& -  &10& $5\times 10^4$\\
                 
                 Case 5-$r=8$& 3+512$\times$5+1& $E_{j,i}^d$& -  &10& $5\times 10^4$\\
                 
                 Case 5-$r=10$& 3+512$\times$5+1& $E_{j,i}^d$& -  &10& $5\times 10^4$\\
                 \hline
	    \end{tabular}}
	\label{table:training details}
\end{table}
\subsection{Electrostatics Field Problems Defined by Poisson Equations}
The Poisson equation is widely used to describe the phenomenon of steady heat conduction, fluid dynamics, and the electrostatic field. Consider case 1, which is an electrostatic field problem defined by the Poisson equation, characterized by corner singularity due to the slit in the domain. The equation is formulated as follows: 
\begin{align}
\begin{cases}
 \nabla^2  u(\boldsymbol{x}) +1=0 ,&\boldsymbol{x} \in (-1,1)\times(-1,1)\backslash[0,1)\times\{0\},
\\
u(\boldsymbol{x})=0,&\boldsymbol{x} \in \partial \Omega,
\end{cases}
\end{align}
where $u(\boldsymbol{x})$ represents the unknown electric potential, the electric field gradient caused by space charge is -1 $\text{V}$ $\text{m}^{-2}$, the boundary potential is set as 0 $\text{V}$, and the computational domain is $[-1,1]\times[-1,1]$ $\text{m}^2$.

\begin{figure}[htbp]
	\centerline{\includegraphics[width=0.75\columnwidth,draft=false]{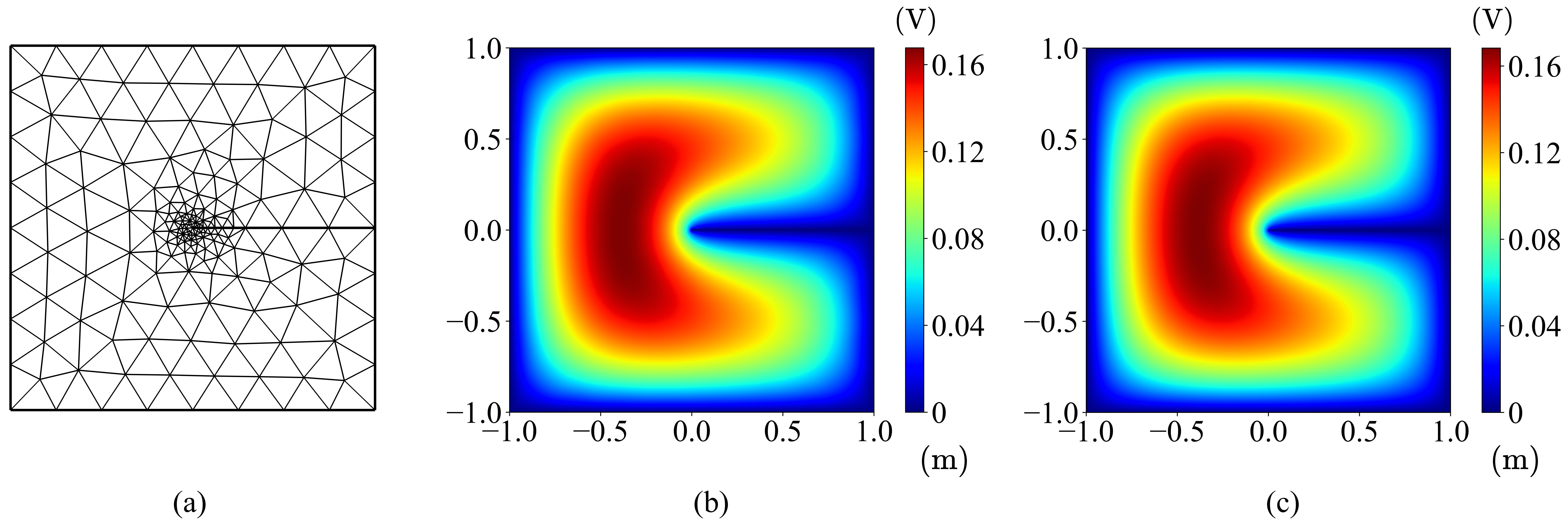}}
	\caption{Case 1: an electrostatic field problem with corner singularity. a) The mesh with local refinement, b) the numerical solution calculated by PD-GraphSAGE, and c) the reference solution calculated by FEM.} \label{fig:corner_singularity_1}
\end{figure}

\begin{figure}[htbp]
	\centerline{\includegraphics[width=1\columnwidth,draft=false]{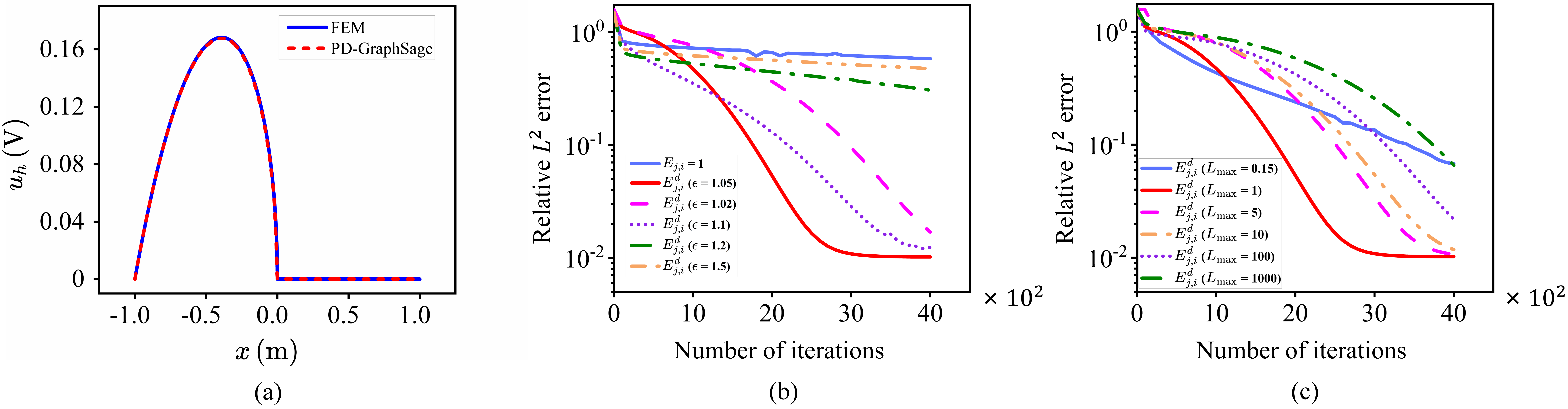}}
	\caption{Singularity area results and training comparison with different parameter settings of case 1. a) The comparison of calculated curves at $y=0$, b) the relative $L^2$ error of training PD-GraphSAGE separately with different $\epsilon$ of $E_{j,i}^d$ and $E_{j,i}=1$, and c) the relative $L^2$ error of training PD-GraphSAGE with different $L_\text{max}$ of $E_{j,i}^d$, where the largest element edge length of mesh is 0.1484 m $\textless$ 0.15 m.} \label{fig:corner_singularity_2}
\end{figure}

Although this PDE lacks the analytical solution, it exhibits an asymptotic behavior at the origin as $u(\boldsymbol{x})=u(r,\theta)\sim r^{\frac{1}{2}}\sin\frac{\theta}{2}$. Models featuring such configurations are extensively used to evaluate adaptive FEMs and adaptive PINN methods.\textsuperscript{\cite{amuthan-a-2021-spinn, e-2018-deep-ritz-method}} In our experiment, we refine the mesh near the origin as depicted in \textbf{Figure \ref{fig:corner_singularity_1}}a to include more nodal points for the evaluation of the loss function. Consequently, a total of 697 evaluated points, comprising 613 collocation points and 84 boundary points with hard enforcement, are gathered from the discrete meshes for this simulation. The reference solution is obtained by FEM based on the refined mesh of 65236 degrees of freedom (DOF). The distance-related edge feature $E_{j, i}^d$ is employed in this case and the scale factor $\delta$ is set to $1\times 10^{-1}$. As a result, the contour of the PD-GraphSAGE solution aligns well with the FEM reference, as depicted in Figure \ref{fig:corner_singularity_1}b and \ref{fig:corner_singularity_1}c. It is noteworthy that the maximum absolute error is $6.56 \times 10^{-3}$, whereas the adaptive PINN method\textsuperscript{\cite{amuthan-a-2021-spinn}} necessitates 710 evaluated points to achieve the level of $2 \times10^{-2}$. Additional details provided in \textbf{Figure \ref{fig:corner_singularity_2}}a show that the solution calculated by PD-GraphSAGE exhibits strong agreement with the FEM solution in the singularity area. Moreover, as demonstrated in Figure \ref{fig:corner_singularity_2}b, the distance-related edge feature $E_{j, i}^d$ ($\epsilon=1.05$) shows the best performance in contributing to the convergent process compared to other edge feature settings. We then investigate the influence of $L_\text{max}$ to $E_{j,i}^d$ when $\epsilon=1.05$, and as the results are shown in Figure \ref{fig:corner_singularity_2}c, $L_\text{max} = 1$ m presents the best convergent speed and accuracy. Therefore, $\epsilon=1.05$ and $L_\text{max} = 1$ m are set as the default in the following cases. Finally, since the solution of PD-GraphSAGE relies on mesh generation, we also study how the number of evaluated points affects the convergence of this method, and the result is shown in \textbf{Table \ref{table:convergence analysis}}: As the mesh becomes finer, the accuracy is higher, but the convergence rate is slower, which meets our expectation since FEM using Galerkin method show the same tendency. 
\begin{table}[htbp]
	\centering
	\caption{Convergence analysis experiments for PD-GraphSAGE of case 1.}
        \renewcommand\arraystretch{1.4}
	\setlength{\tabcolsep}{4mm}{
		\begin{tabular}{ccc}
			\hline
			  Evaluated points&Relative $L^2$ error&Convergent epochs\\
                \hline
			  697&$1.02\times 10^{-2}$&$4\times 10^3$\\
                1077&$7.68\times 10^{-3}$&$4.8\times 10^3$\\
                2172& $6.37\times 10^{-3}$&$2.08\times 10^4$\\
                \hline
	\end{tabular}}
	\label{table:convergence analysis}
\end{table}

\begin{figure}[htbp]
	\centerline{\includegraphics[width=0.75\columnwidth,draft=false]{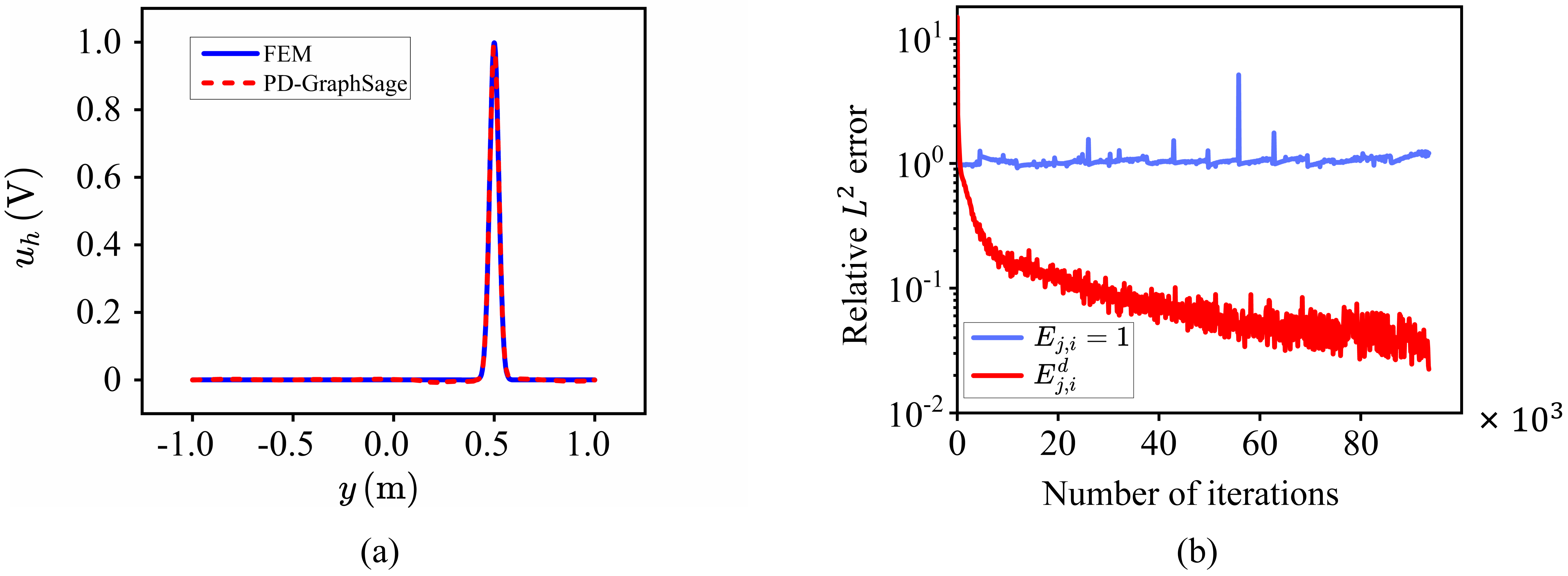}}
	\caption{Singularity area results and training comparison with different edge feature strategies of case 2. a) The comparison of calculated curves at $x=0.5$ and b) the relative $L^2$ error of training PD-GraphSAGE separately with $E_{j,i}^d$ and $E_{j,i}=1$.} \label{fig:peak_singularity_2}
\end{figure}

Consider case 2, which is an electrostatic field problem defined by the Poisson equation, with singularity caused by a peak source term. The equation can be formulated as: 
\begin{align}
\begin{cases}
 \nabla^2  u(\boldsymbol{x}) =f(\boldsymbol{x}) ,&\boldsymbol{x} \in (-1,1)\times(-1,1),
\\
u(\boldsymbol{x})=g(\boldsymbol{x}),&\boldsymbol{x} \in \partial \Omega.
\end{cases}
\end{align}
The source term is specially designed to create the singularity, and the analytical solution is given as: 
\begin{align}
	u(\boldsymbol{x}) = e^{-1000\left( (x-\frac 1 2)^2+(y-\frac 1 2)^2\right)},
\end{align}
\begin{figure}[htbp]
	\centerline{\includegraphics[width=0.75\columnwidth,draft=false]{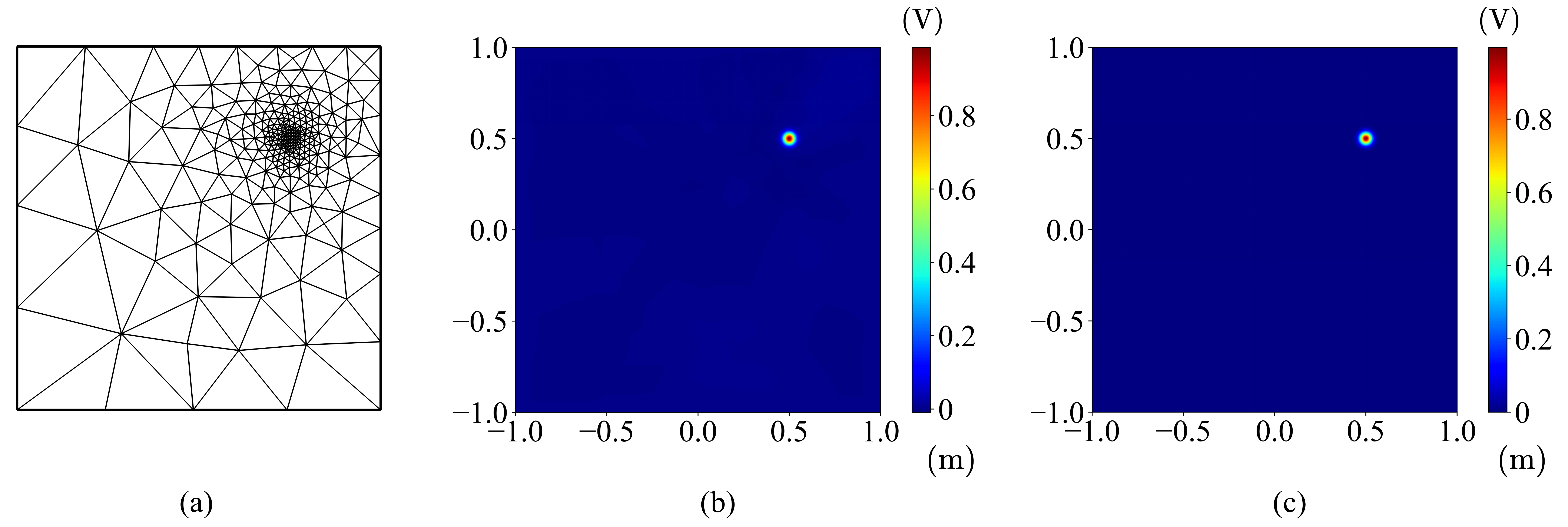}}
	\caption{Case 2: an electrostatic field problem with a peak source. a) The mesh with local refinement, b) the numerical solution calculated by PD-GraphSAGE, and c) the reference solution calculated by FEM.} \label{fig:peak_singularity_1}
\end{figure}
which peaks at (0.5 m, 0.5 m) and decreases rapidly away from this point. Several adaptive sampling strategies have been investigated for this scenario.\textsuperscript{\cite{tang-ke-jun-2023-das-pinn,gao-zhi-wei-2023-fpinn}} In our experiment, as shown in \textbf{Figure \ref{fig:peak_singularity_1}}a, the mesh is refined around the peak area to address the singularity instead of using adaptive sampling, since the accuracy is ensured in our framework. Accordingly, a total of 1081 evaluated points, including 1033 collocation points and 48 boundary points with hard enforcement are collected for training. The distance-related edge feature $E_{j, i}^d$ is also employed in this experiment to enhance convergence. The reference solution obtained by FEM is based on the refined mesh of 50225 DOF. As a result, as demonstrated in Figure \ref{fig:peak_singularity_1}b and \ref{fig:peak_singularity_1}c, the PD-GraphSAGE trained solution of the field is consistent with the FEM reference. The relative $L^2$ error of the PD-GraphSAGE solution is $2.25\times 10^{-2}$, while the adaptive PINN method\textsuperscript{\cite{gao-zhi-wei-2023-fpinn}} achieves $9.04\times 10^{-2}$ with more than 2000 evaluated points. As further detailed in \textbf{Figure \ref{fig:peak_singularity_2}}, in the singularity area, the solution provided by PD-GraphSAGE agrees well with the FEM solution, and the PD-GraphSAGE fails to converge when the edge feature is $E_{j, i}=1$, while it performs well with $E_{j, i}^d$ .

Based on the results of case 1 and case 2, it can be concluded that PD-GraphSAGE with distance-related edge feature $E_{j, i}^d$ demonstrates the capability to tackle singularity problems effectively.

\subsection{Electromagnetic Problems Defined by Helmholtz Equations}
\begin{figure}[htbp]
	\centerline{\includegraphics[width=0.6\columnwidth,draft=false]{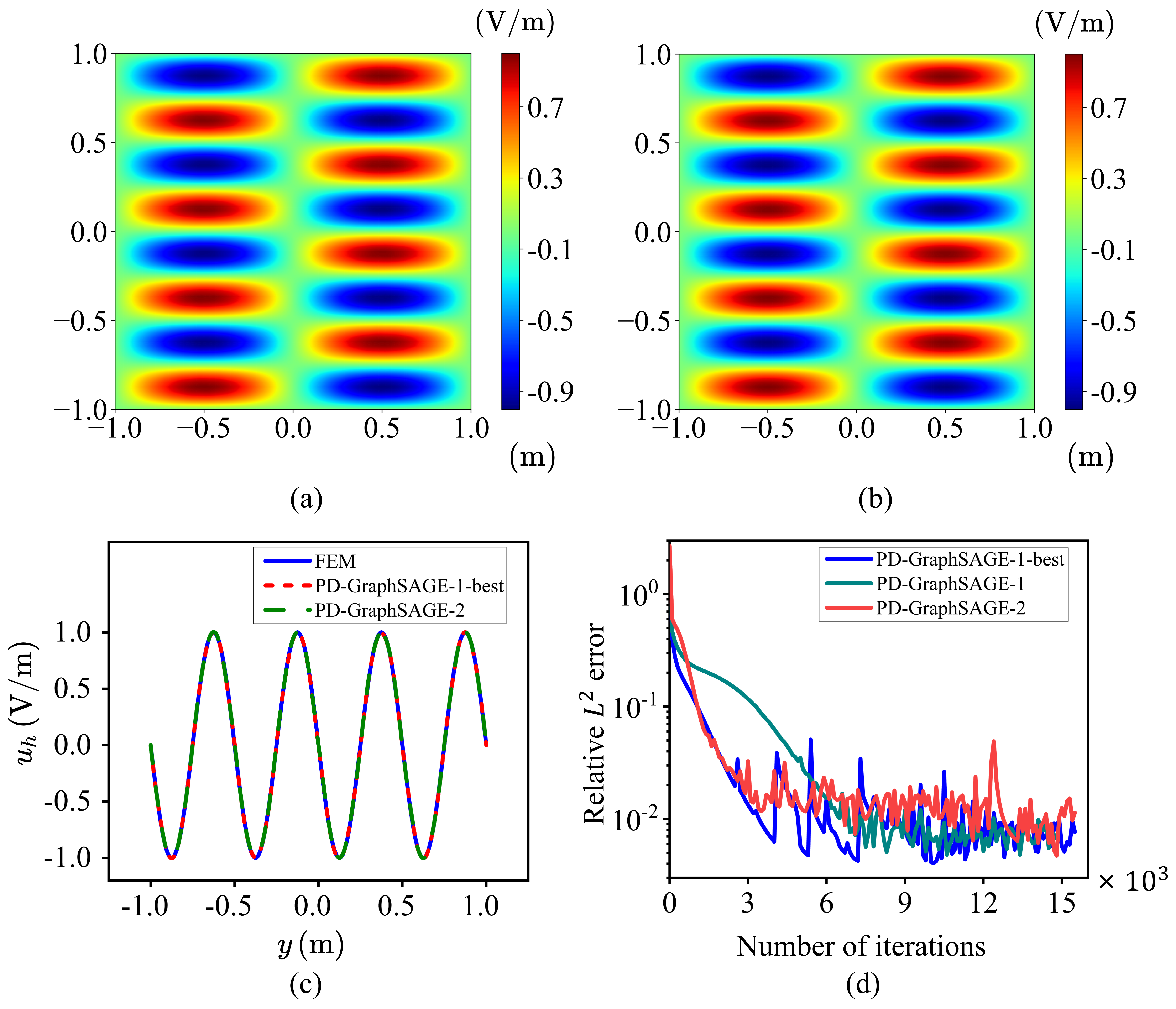}}
	\caption{Case 3: an electromagnetic problem with an oscillating source. a) The numerical solution calculated by PD-GraphSAGE-2, b) the reference solution calculated by FEM, c) the comparison of calculated curves at $x=0.5$, and d) the relative $L^2$ error of training PD-GraphSAGE with different feature mapping strategies. } \label{fig:general_oscillation}
\end{figure}

The Helmholtz equation is commonly applied to describe the behavior of acoustics, quantum mechanics, and electromagnetism. In this study, case 3 is examined: an electromagnetic problem defined by the Helmholtz equation, characterized by an oscillating source. The equation is formulated as:
\begin{align}
	\begin{cases}
		\nabla^2 u(\boldsymbol{x}) -u(\boldsymbol{x}) = f(\boldsymbol{x}) ,&\boldsymbol{x} \in (-1,1)\times(-1,1),\\
		u(\boldsymbol{x})=0,&\boldsymbol{x} \in \partial \Omega,
	\end{cases}
\end{align}
where $u(\boldsymbol{x})$ denotes the unknown electric field intensity component, the corresponding boundary electric field intensity is set as 0 $\text{V/m}$, the computational domain is $[-1,1]\times[-1,1]$ $\text{m}^2$, and the imposed electric field intensity density $f(\boldsymbol{x})$ is provided as:
\begin{align}
	f(\boldsymbol{x}) = -(\alpha_1^2 +\alpha_2^2+1) \pi^2 \sin(\alpha_1 \pi x)\sin(\alpha_2 \pi y).
\end{align}
To highlight the advantages of our framework, $\alpha_1=1$ and $\alpha_2=4$ are selected for convenient comparison with other methods. In our experiment, the computational domain is uniformly discretized due to the even distribution of the oscillatory behavior, leading to a total of 1933 evaluated points being collected, including 1781 collocation points and 152 boundary condition points with hard enforcement. The reference solution is obtained via FEM with a refined mesh of 50225 DOF.  The distance-related edge feature $E_{j,i}^d$ is omitted as it could lead to deteriorating conditions when the local solution fluctuates rapidly. Instead, the original Fourier feature mapping $\gamma_1(\mathbf f)$ and the proposed feature mapping $\gamma_2(\mathbf f)$ were utilized for comparison. For simplicity, we refer to the combination of PD-GraphSAGE with $\gamma_1(\mathbf{f})$ and $\gamma_2(\mathbf{f})$ as PD-GraphSAGE-1 and PD-GraphSAGE-2, respectively. PD-GraphSAGE-1-best means the setting that achieves the best performance in convergent speed and accuracy when using the feature mapping strategy $\gamma_1(\mathbf{f})$, and PD-GraphSAGE-1 has the same hyperparameters as PD-GraphSAGE-2 for the convenience of direct comparison. Consequently, as depicted in \textbf{Figure \ref{fig:general_oscillation}}a and \ref{fig:general_oscillation}b, the field solution obtained by PD-GraphSAGE-2 is accordant with the FEM reference, with a relative $L^2$ error of $4.72\times 10^{-3}$. It is worth noting that the adaptive PINN method\textsuperscript{\cite{levi-mcclenny-2022-self-adaptive-pinn-soft-attention}} needed more than $1\times 10^5$ evaluated points to achieve the same level of accuracy. In Figure \ref{fig:general_oscillation}c, solutions of PD-GraphSAGE-1-best and PD-GraphSAGE-2 are both coherent well with the FEM reference. As observed in Figure \ref{fig:general_oscillation}d, PD-GraphSAGE-1-best converges slightly faster than PD-GraphSAGE-2, while all ultimately reached a similar error level with only a marginal difference. 

Consider case 4, which is an electromagnetic problem defined by the Helmholtz equation with a high wave number $k$:
\begin{align}
	\begin{cases}
		\nabla^2 u(\boldsymbol{x}) +k^2u(\boldsymbol{x}) = 0 ,&\boldsymbol{x} \in (0,1)\times(0,1), \\
		u(\boldsymbol{x})=\text{sin}(k\boldsymbol{x}),&\boldsymbol{x} \in \partial \Omega,
	\end{cases}
\end{align}
where the computational domain is $[0,1]\times[0,1]$ $\text{m}^2$ and the boundary electric field intensity is defined as a sine function of $k$. The quality of the discrete numerical solution for the Helmholtz equation is closely related to $k$, and the oscillation phenomenon intensifies as $k$ increases. 

\begin{figure}[htbp]
	\centerline{\includegraphics[width=0.6\columnwidth,draft=false]{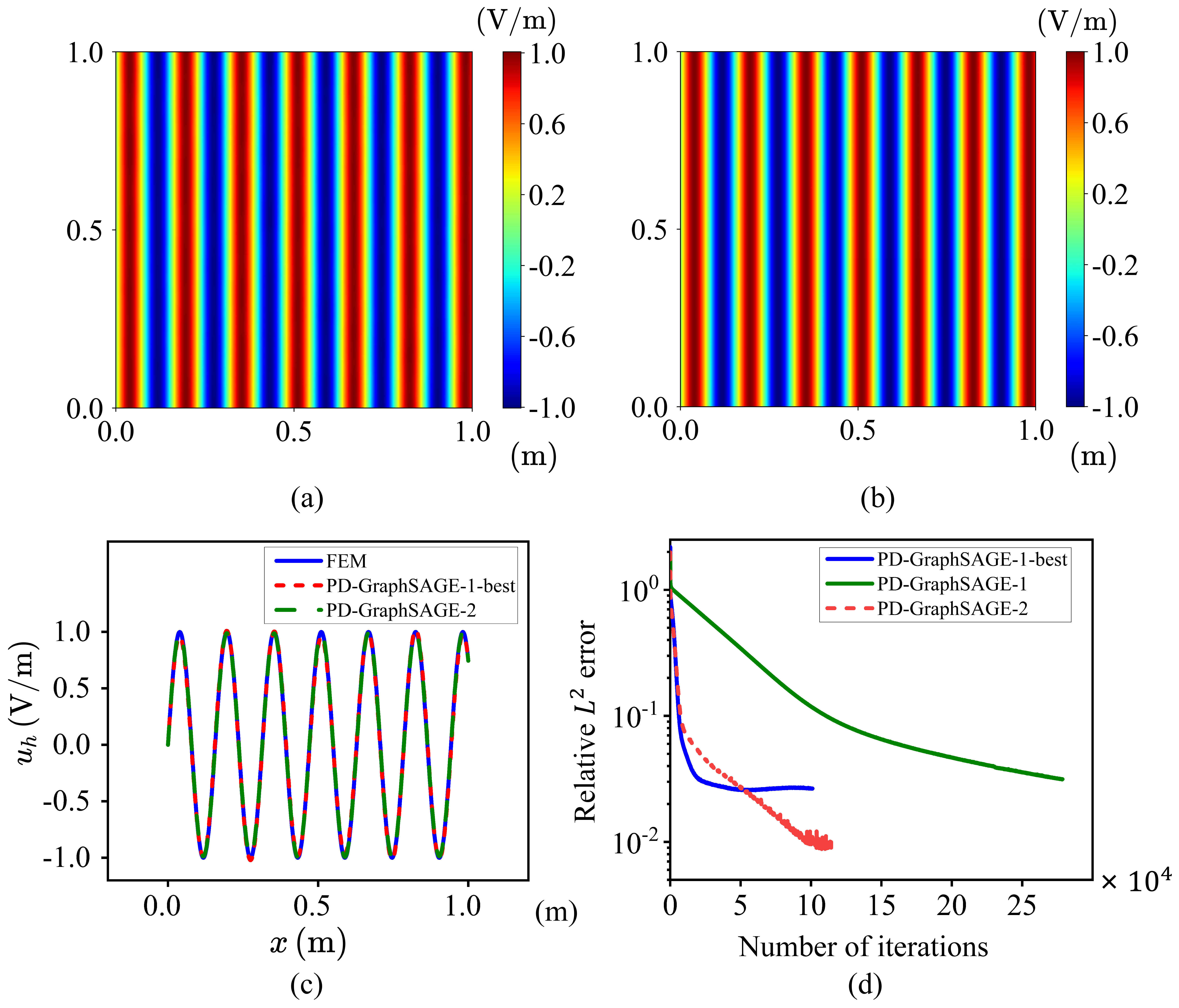}}
	\caption{Case 4: an electromagnetic problem with a high wave number $k$. a) The numerical solution calculated by PD-GraphSAGE-2, b) the reference solution calculated by FEM, c) the comparison of calculated curves at $y=0.5$, and d) the relative $L^2$ error of training PD-GraphSAGE with different feature mapping strategies.} \label{fig:wave_number_oscillation}
\end{figure}

In our experiment, $k$ is set to 40 $\text{m}^{-1}$. The computational domain undergoes uniform discretization, yielding 12765 evaluated points for training, including 12365 collocation points and 400 boundary points with hard enforcement. The reference solution is calculated using FEM on a refined mesh of 50225 DOF. Similar to case 3, the edge feature $E_{j,i}^d$ is omitted. Both the original Fourier feature mapping $\gamma_1(\mathbf f)$ and the proposed feature mapping $\gamma_2(\mathbf f)$ are employed for further comparison. As a result, shown in \textbf{Figure \ref{fig:wave_number_oscillation}}a and \ref{fig:wave_number_oscillation}b, the field solution obtained by PD-GraphSAGE-2 is consistent with the FEM reference, exhibiting a relative $L^2$ error of $8.79\times10^{-3}$. As noted by Fang,\textsuperscript{\cite{fang-2020-pinn-metamaterial}} when the wave number $k$ is relatively high (exceeding 23), neural networks face challenges in accurately approximating the true solution. However, our method maintains relatively acceptable accuracy even at a high wave number of 40. In Figure \ref{fig:wave_number_oscillation}c, it is evident that the solution achieved by PD-GraphSAGE-1-best is slightly less accurate in the peak area than that of PD-GraphSAGE-2. A more detailed comparison, as illustrated in Figure \ref{fig:wave_number_oscillation}d, reveals that PD-GraphSAGE-1-best converges more rapidly than PD-GraphSAGE-2, while PD-GraphSAGE-2 attains a lower ultimate error, thereby highlighting the effectiveness of the proposed feature mapping strategy $\gamma_2(\mathbf f)$.

Drawing upon the findings from case 3 and case 4, it is evident that PD-GraphSAGE, when employing the proposed feature mapping strategy $\gamma_2(\mathbf f)$, exhibits a robust ability to address oscillating problems, sometimes outperforming those when combined with the Fourier feature mapping $\gamma_1(\mathbf f)$.

\subsection{Steady heat conduction Problems Defined by Parametric Poisson Equations}
Consider case 5, a heat conduction problem defined by the Poisson equation and parameterized by the source term. The equation is formulated as follows:
\begin{align}
\begin{cases}
c\nabla^2 u(\boldsymbol{x})  = f(\boldsymbol{x}) ,&\boldsymbol{x} \in (0,1)\times(   0,1),\\
u(\boldsymbol{x})=0,&\boldsymbol{x} \in \partial \Omega,
\end{cases}
\end{align}
where $u(\boldsymbol{x})$ is the unknown temperature, the thermal parameter $c$ is 1 $\text{W}$ $\text{m}^{-1}$ $\text{K}^{-1}$, the boundary temperature is 0 $\text{K}$, and $f(\boldsymbol{x})$ is the source function generated from the Gaussian random field,\textsuperscript{\cite{bruno-2017-2d-gaussian-random-fields}} which is scale-invariant and two-dimensional correlated. The correlations are characterized by a scale-free spectrum $\text{P(k)}\sim\frac{1}{k^\frac{1}{r}}$, where $r$ indicates the strength of correlation and $f(\boldsymbol{x})$ gets smoother as $r$ increases.

\begin{figure}[htbp]
	\centerline{\includegraphics[width=0.75\columnwidth,draft=false]{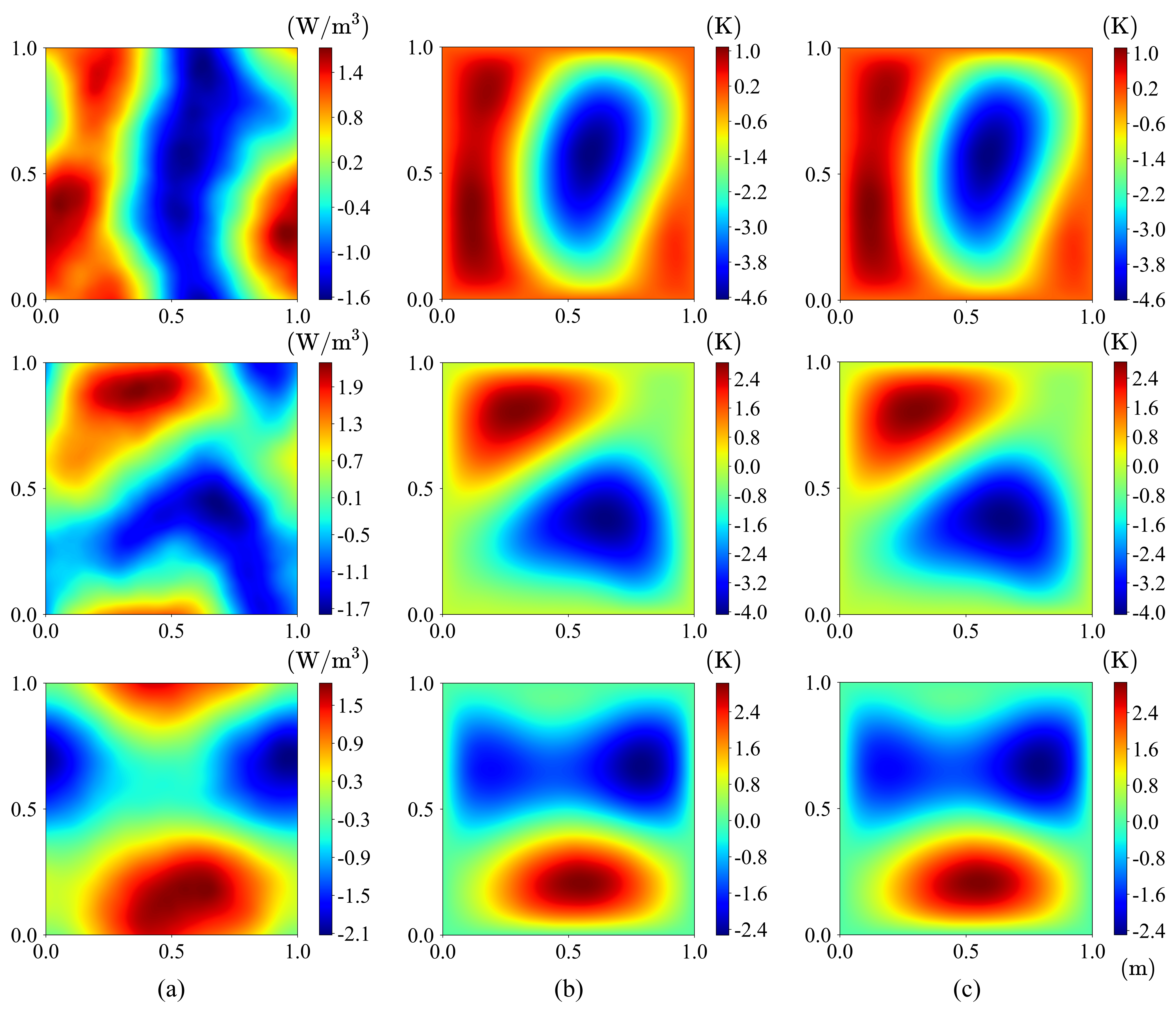}}
	\caption{Three test results (3 rows) for well-trained parametric heat conduction surrogate model when $r$ = 6. a) Source distribution, b) inference solution calculated by PD-GraphSAGE, and c) reference solution calculated by FEM.}\label{fig:test r = 6}
\end{figure}

\begin{figure}\label{fig:r=8,10}
    \centering
    \includegraphics[width=0.75\linewidth]{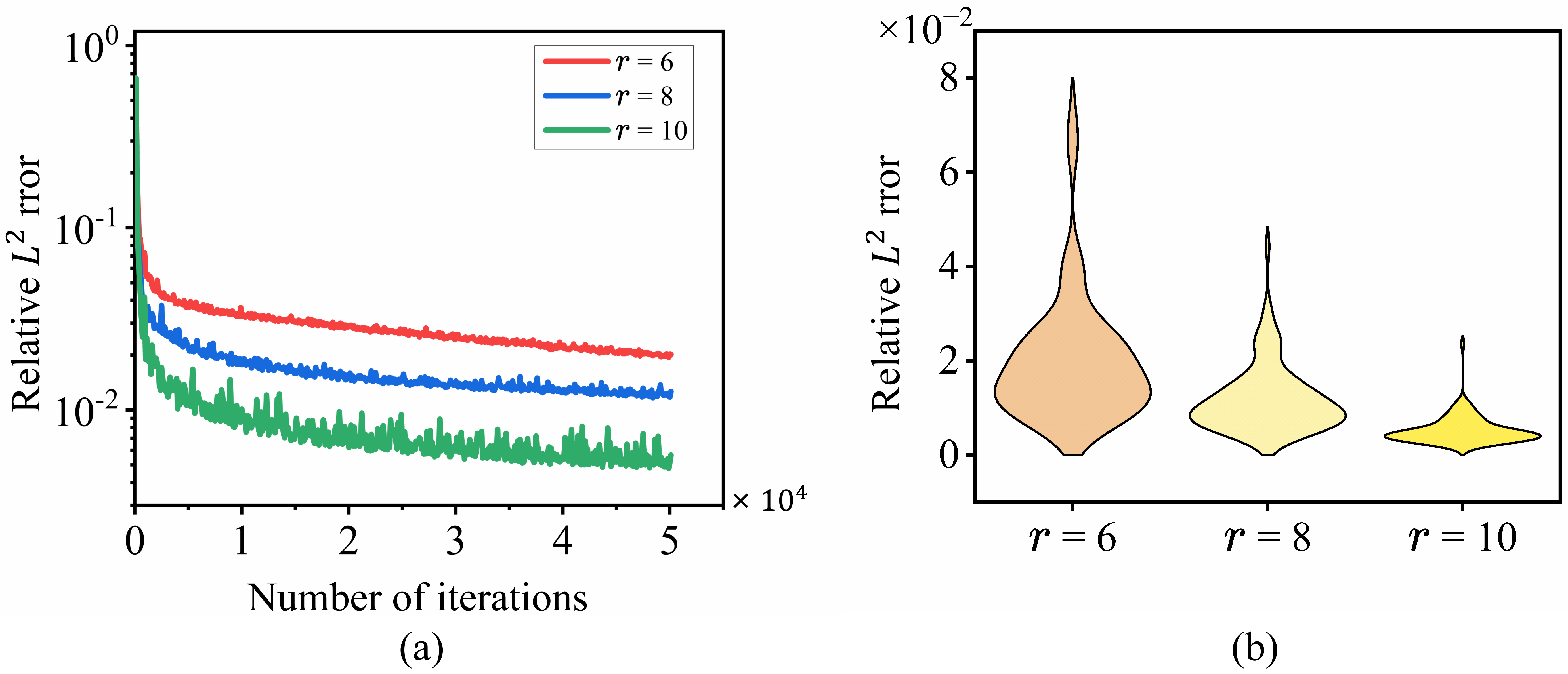}
    \caption{Comparison for different coefficients $r$ of case 5. a) The relative $L^2$ error of training PD-GraphSAGE with different $r$ and b) the relative $L^2$ error distribution on 100 random test samples for different $r$.}
    \label{fig:relative-violin}
\end{figure}

In our experiment, the coefficient $r$ is firstly set to 6, and the computational domain is uniformly discretized, resulting in a total of 157 evaluated points, including 117 collocation points and 40 boundary points with hard enforcement in each input graph. The FEM reference solution is based on the same mesh as PD-GraphSAGE. The random heat source vector $\boldsymbol{f(\mathbf{x})}\in\mathbb{R}^{157}$, sampled on the corresponding nodal locations $\mathbf{x}$ of the generated heat source distribution, is considered as the parameter vector $\boldsymbol{\mu}$, namely one of the nodal features for training. A total of $1\times 10^{3}$ Gaussian random heat sources are produced for training, and another 50 are for validation. Moreover, the edge feature $E_{j,i}^d$ and Batch Normalization\textsuperscript{\cite{Ioffe-2015-batchnormalization}} are utilized to promote convergence. After training for $5\times 10^4$ epochs, 3 new heat source samples (not included in the training and validating datasets) are generated to test the model. As shown in \textbf{Figure \ref{fig:test r = 6}}, the inference solutions from the parametric heat conduction surrogate model closely match the FEM references for all 3 test samples. To further quantify the effectiveness of the trained model, another 100 test input graphs are constructed, and the average relative $L^2$ error is $1.99\times 10^{-2}$, which is an acceptable accuracy for fast physical process simulations. Furthermore, for comparison, we also train a data-driven GraphSAGE under the same setting and datasets. This data-driven model requires $5.1\times 10^3$ epochs to converge and reach the average relative $L^2$ error of  $1.37\times 10^{-2}$. However, it necessitates the preparation of a substantial amount of data in advance. From another perspective, even a well-trained data-driven model fails to provide a highly accurate solution, highlighting the significant challenges associated with this problem. Yet, as the size of the training datasets increases, an improvement in accuracy is anticipated for both the data-driven model and the physics-driven one, but the required training time will also grow. Back to PD-GraphSAGE, to study the influence of the coefficient $r$ on the model and the robustness of the model, the experiment configuration is kept unchanged, and models with $r=8$ and 10 are retrained. \textbf{Figure \ref{fig:relative-violin}}a illustrates that the relative $L^2$ error decreases as $r$ increases. In Figure \ref{fig:relative-violin}b, the statistical relative $L^2$ error distributions of 100 test samples for $r=6,8,10$ are present respectively, clearly showcasing that the relative $L^2$ error of most test samples is stably distributed under a low level. Finally, as demonstrated in \textbf{Table \ref{table:compare with FEM, PINN}}, the computing speed for the well-trained parametric heat conduction surrogate model is approximately $4.2\times$ than that for FEM, and hundreds of times than that of PD-GraphSAGE. 
\begin{table}[htbp]
	\centering
	\caption{Solving time (seconds) when $r=6$ for PD-GraphSAGE (same configuration with case 1), FEM, and the well-trained parametric heat conduction surrogate model.}
        \renewcommand\arraystretch{1.4}
	\setlength{\tabcolsep}{4mm}{
		\begin{tabular}{cccc}
			\hline
			  Hardware platform&PD-GraphSAGE&FEM& Parametric heat conduction surrogate model\\
                \hline
			  AMD Ryzen 5 5600x&-&0.063&   0.015\\
                RTX 3090 Ti&14.183&-&-\\
                \hline
	\end{tabular}}
	\label{table:compare with FEM, PINN}
\end{table}

Based on case 5, it can be concluded that the parametric heat conduction surrogate model operates more rapidly than FEM, offering acceptable accuracy and excellent robustness. This is expected to accelerate diverse physical process simulations. Furthermore, as demonstrated by cases 1 to 4, PD-GraphSAGE utilizing the proposed edge feature $E_{j,i}^d$ and the feature mapping strategy $\gamma_2(\mathbf f)$ successfully addresses challenging problems featuring irregular solutions, highlighting the wide applicability of our method. 

\section{Conclusion}
In this work, we present a physics-driven GraphSAGE method to address problems in computational physics governed by PDEs with irregular solutions and to construct the parametric PDE surrogate model for rapid solution inference. The proposed PD-GraphSAGE employs the Galerkin method and piecewise polynomial nodal basis functions to simplify the approximation of PDE solutions. The Graph is utilized to represent the physical domain, which is computationally attractive due to reducing the need for evaluated points. By combining the proposed distance-related edge feature $E_{j,i}^d$ and the feature mapping strategy $\gamma_2(\mathbf f)$, PD-GraphSAGE can accurately capture the singular and oscillatory solution of PDEs-governed physical problems. Furthermore, we have successfully established the parametric PDE surrogate model using our proposed method, which proves to be several times faster than FEM and is anticipated to be applicable in diverse real-time scenarios, such as temperature analysis for devices and electrothermal monitoring.

Moreover, GraphSAGE employs the full neighbor sampling method in this study, and the graph size is not sufficiently large. As a result, the capability of GraphSAGE for large-scale graph learning is not fully exploited. Additionally, given that our experiments were conducted with two-dimensional linear PDEs within a regular computational domain, expansion to more complex and realistic problems governed by three-dimensional nonlinear PDEs and irregular computational domains is feasible due to the general principles of the Galerkin method. Besides, other parameters, such as material parameters and boundary conditions, are easily added as the parameter vector $\boldsymbol{\mu}$ for physics-driven training. In addition, the inference speed of the surrogate model has surpassed that of FEM on the same order of magnitude, which is currently limited by the experimental device. It can be easily inferred that this speed advantage will be more prominent as the physical model becomes more complex because the inference time of the surrogate model is approximately equal to the time of one forward propagation after being well-trained, yet the FEM calculation time depends on the complexity of the model. Finally, the physical system defined by spatiotemporal PDEs and adaptive mesh refinement PD-GraphSAGE are also promising fields to extend to. 

Overall, there are compelling reasons to believe that the proposed framework can be widely applied across various physical process simulations owing to its high accuracy, fast speed, flexible scalability, and strong robustness.

\end{document}